\documentclass{article}

\usepackage[utf8]{inputenc}
\usepackage[T1]{fontenc}
\usepackage{url}
\usepackage{booktabs}
\usepackage{amsfonts}
\usepackage{amsmath}
\usepackage{amssymb}
\usepackage{xcolor}
\usepackage{graphicx}
\usepackage{multirow}
\usepackage{enumitem}
\usepackage{array}
\usepackage{natbib}
\usepackage[margin=1in]{geometry}
\usepackage{upquote}
\usepackage{hyperref}
\usepackage{float}

\title{Post-Training on Office Work Improves Software Engineering: A Behavioral Account of Cross-Domain Transfer}

\author{
  Logan Ritchie\thanks{Correspondence to: \texttt{research@surgehq.ai}},
  Sushant Mehta, Liudas Panavas, and Edwin Chen\\
  Surge AI
}

\date{}

\begin{document}

\maketitle

\begin{abstract}
Long-horizon tasks require agents to maintain coherent state and goals across nested and branching work. We call this capability \emph{goal-directed execution} (GDE): the repeated application of four behaviors, namely selecting goals, constructing task-relevant state, maintaining fidelity to higher-level objectives, and verifying completion against the environment. We hypothesize that long-horizon post-training strengthens these behaviors across domains. We test this by post-training Qwen3.5-122B-A10B on 363 Long-Horizon Multi-Tool Agent (LHMTA) tasks drawn from office workflows. The collection contained no software-engineering tasks, yet the model's pass@1 improved by 5.8 points on SWE-Bench Pro. Matched trajectory analysis shows gains in all four GDE behaviors in both office workflows and software repositories. Aggregate SWE-Bench Pro statistics showed related changes in information gathering, implementation, and verification. Together, the results support a behavioral interpretation in which long-horizon post-training changed how the model organized and applied knowledge across tasks, with effects extending beyond the training domain.
\end{abstract}

\section{Introduction}
\label{sec:intro}

Post-training is often described in terms of the content and domain of its examples. However, long-horizon tasks may also exercise broader ways of organizing behavior. Across a complex task, an agent must decide what to achieve next, maintain an accurate view of the environment, preserve higher-level requirements during local work, and determine whether its actions succeeded. These demands can recur even when the tools and subject matter change.

We use \emph{goal-directed execution} (GDE) to describe the agent's ability to stay on track: maintaining coherent goals and task-relevant state across nested and branching work, and using them to select, coordinate, and evaluate behavior over time. We operationalize GDE through four behavioral capabilities:

\begin{enumerate}
  \item \textbf{Goal formation:} deriving the correct immediate target from the parent goal and current working state.
  \item \textbf{State construction:} gathering, interpreting, integrating, and preserving the environment information needed to guide local and higher-level decisions.
  \item \textbf{Goal stability:} maintaining higher-level requirements and intended outcomes while pursuing lower-level work.
  \item \textbf{Verification:} determining what evidence would establish that the intended state has been reached and obtaining that evidence at intermediate and final boundaries.
\end{enumerate}

We represent task execution as a goal loop. At any level, the agent forms a goal, acts on or queries the environment, updates its working state from the result, and checks that state against the goal. An action that is too abstract to execute may decompose into lower-level goals, so complex tasks contain nested and branching goal loops. GDE is the capability required to execute these loops and coordinate their goals and state updates across levels.

We hypothesize that GDE is a domain-general requirement for complex agentic tasks. Training that strengthens this capability in one domain may therefore improve performance in others. We examine this using a domain-mismatched source and target. We post-trained Qwen3.5-122B-A10B on 363 LHMTA tasks: realistic workflows over documents, spreadsheets, web research, planning, file manipulation, and office services, exposed as RL environments through Model Context Protocol tools \citep{anthropic2024mcp}. The collection contained no software-engineering tasks. The trained model nevertheless improved by 5.8 percentage points on SWE-Bench Pro \citep{deng2025swebenchpro}.

The training tasks supplied no repository conventions, parser semantics, or code-level solutions. To examine what changed, we compared matched base and trained trajectories. The trained model more reliably formed local goals that served the broader task, built and maintained relevant working state, preserved parent requirements during lower-level work, and verified substantive completion conditions. The same differences appeared in office workflows and software repositories.

Aggregate SWE-Bench Pro statistics provided broader but indirect evidence. With a similar volume of retrieval, the trained model repeated less of what it read; its patches overlapped more with the reference implementation while adding substantially fewer lines; and the share of trajectories running a formal test nearly doubled. Goal stability had no clean aggregate proxy, so evidence for it came from the matched trajectory comparisons.

This paper makes three contributions. First, it documents cross-domain transfer from non-software long-horizon post-training to SWE-Bench Pro. Second, it operationalizes GDE in learned agents through a recursive goal-loop representation and four behavioral capabilities. Third, it provides paired trajectory evidence and aggregate behavioral measures consistent with these capabilities improving after post-training. GDE provides an explanatory account of the observed transfer, but the proposed causal link to long-horizon task structure remains a hypothesis.

\section{Related Work}
\label{sec:related}

\paragraph{Agent architectures.}
Research on language agents has concentrated on how to build them: ReAct interleaves reasoning with actions \citep{yao2023react}, Reflexion adds verbal self-reflection \citep{shinn2023reflexion}, search wraps inference-time exploration around the model \citep{yao2023tot, zhou2024lats}, and decomposition-based prompting reduces problems to simpler subproblems \citep{zhou2022least, prasad2024adapt}. Recent long-horizon scaffolds connect decomposition to context management more directly. HiAgent organizes working memory into subgoal-based chunks, replacing completed branches with summaries while allowing their details to be retrieved \citep{hu2025hiagent}. ReCAP recursively decomposes plans and reintroduces the parent plan after subtask execution so that new observations can update higher-level work \citep{zhang2025recap}. Both prescribe structures for improving inference-time behavior; we use related concepts to describe behavior learned through post-training. The pattern predates language models: hierarchical RL builds subtask hierarchies into the agent as learnable structure \citep{sutton1999options, dietterich2000maxq}, and HTN planning decomposes tasks through hand-authored methods \citep{ghallab2004automated}. Cognitive architectures provide the closest precedents for the full model: classically, Soar spawns a subgoal automatically whenever an impasse blocks progress, over a working memory representing the current situation \citep{laird1987soar}; more recently, CoALA carries the same concepts to language agents, organizing them around a working memory hub with distinct internal and external actions \citep{sumers2024coala}. All of these contributions answer an engineering question: what mechanism or scaffolding makes an agent act better? They coexist with a broader trend toward general methods that scale with computation rather than task-specific structure \citep{sutton2019bitter}. We argue that structured accounts of task completion still hold value for interpretation and evaluation: a learned agent must still form goals, build state, and verify outcomes, whether or not anything inside it resembles a goal stack, so the classical vocabulary remains useful for describing agent behavior.

\paragraph{Cognitive foundations.}
The goal-loop representation is functionally aligned with the test-operate-test-exit (TOTE) framework of \citet{miller1960plans}. Both describe behavior as recursive, hierarchical cycles that compare an intended condition with represented current state, operate to reduce the discrepancy, and test again. The working state resembles their \emph{Image}: accumulated, organized knowledge about the organism and its environment. Their broader account of Plans and Images also includes plan formation, working memory, coordination among plans, and knowledge updated through action.

We operationalize this structure for learned tool-using agents through four behaviorally observable capabilities: goal formation, state construction, goal stability, and verification. Goals, actions, working state, and verification are inferred from reasoning, tool calls, environment feedback, edits, and tests rather than treated as a literal internal architecture. This makes the structure useful for auditing where real agent trajectories diverge and for comparing behavior before and after post-training. The decomposition structure also echoes problem spaces \citep{newell1972human}, GOMS \citep{card1983psychology}, and hierarchical task analysis \citep{annett1967task}. BDI architectures likewise made belief state and the stability of intentions explicit \citep{bratman1987intention, rao1995bdi}. Cognitive control research emphasizes the active maintenance of goals against interference \citep{miller2001integrative, duncan2012goalneglect}.

\paragraph{Behavioral analysis of agent failures.}
Our method characterizes agents entirely by the structure of their behavior, following the behavioral study of machines proposed by \citet{rahwan2019machine}, the treatment of GPT-3 as a participant in cognitive-psychology experiments \citep{binz2023cogpsych}, and machine psychology, which defines itself by analyzing input-output relationships rather than inner workings \citep{hagendorff2024machine}. We extend this paradigm from single-turn probes to long-horizon trajectories. A growing literature catalogs agent failures: repetitive loops \citep{yao2023react}, degradation and non-recovery in multi-turn settings \citep{laban2025lost}, planning shortfalls \citep{valmeekam2023planbench, kambhampati2024llmmodulo}, and goal misgeneralization, in which an agent retains its capabilities yet pursues the wrong goal \citep{langosco2022goal, shah2022goal}. Goal drift has also been measured behaviorally by exposing agents to competing objectives or requiring them to return from a temporary goal to the original one \citep{arike2025goaldrift}. That work studies the stability of a prompt-specified top-level objective; our account also considers higher-level requirements that must remain stable across nested work. Each failure mode is typically named and measured within a single task family, which makes it difficult to say whether two differently named failures in different domains are the same breakdown. Prior work from our group derived a hierarchy of agentic capability levels from failure analysis of frontier models on a realistic RL environment \citep{ritchie2026hierarchy}; here we describe general agent failures as breakdowns at specific locations within a nested goal-loop representation of tasks.

\paragraph{What makes training data valuable.}
Data-selection research has shown that value does not scale straightforwardly with volume: small curated sets can match or beat full corpora \citep{zhou2023lima, xia2024less, liu2024deita}, generalization scales with the number and variety of distinct tasks rather than with examples per task \citep{wei2021flan, chung2022scaling, zhang2024instruction}, and the RL analogue holds for procedurally diverse environments \citep{cobbe2020procgen}. In agentic RL, training domains with richer observations and longer policy trajectories have also been associated with stronger cross-domain retention than domains with greater realism or surface similarity \citep{liu2026generalizationtax}. That analysis covered four environments and intervened only on observation richness, so the role of planning complexity remained correlational. Two results from our own group point the same way: training a small model on roughly a thousand expert-written constraint rubrics improved instruction following on benchmarks it never saw \citep{surge2026complexif}, and training a different base model on CoreCraft, a high-fidelity enterprise environment, improved out-of-distribution tool-use benchmarks \citep{mehta2026corecraft}. Our analysis extends this question to cross-domain transfer: Section~\ref{sec:taskdesign} identifies structural demands that may help explain why behavior learned in one domain remains useful in another.

\section{Goal-Directed Execution in Long-Horizon Agents}
\label{sec:taskmodel}

We use the goal loop to represent task execution and GDE to describe the capability required to run and coordinate these loops reliably. This is a behavioral model: it describes what the agent appears to pursue, what it does, what evidence becomes available, and how later behavior responds. It does not assume that the model contains a literal symbolic goal stack or state register.

A goal loop contains three basic elements:

\begin{itemize}
  \item \textbf{Goal:} the desired state of the environment, including the agent's knowledge of that environment.
  \item \textbf{Action:} an interaction intended to move the environment toward that state.
  \item \textbf{Working state:} the agent's current, behaviorally inferred representation of the environment as it relates to its active and parent goals.
\end{itemize}

Environment state is what is actually true; working state is the agent's current representation of it. The environment is broad enough to include what the agent knows, so an investigation can have a goal such as learning which files are relevant. At the top level, the goal is formed from the task prompt. Within the task, a local goal is the immediate desired state formed from a parent goal and the current working state.

\begin{figure}[H]
\centering
\includegraphics[width=0.75\linewidth]{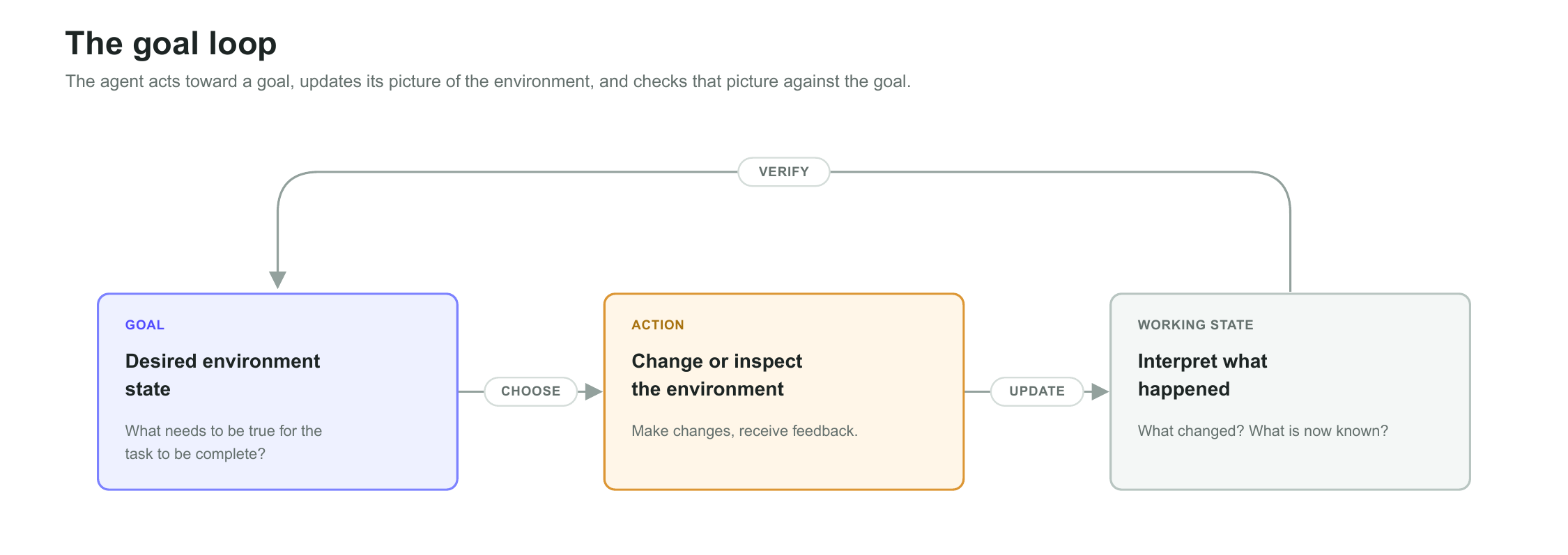}
\caption{The goal loop. The agent forms a goal, acts on or queries the environment, updates its working state from the result, and compares the updated state with the goal; verification connects the working state back to the goal.}
\label{fig:goalloop}
\end{figure}

The agent compares its working state with the goal, chooses an action, observes the result, and updates its working state (Figure~\ref{fig:goalloop}). If the updated state satisfies the goal, the loop exits or returns control to its parent; otherwise, the updated state informs the next action. Because the agent rarely begins with complete knowledge, actions can both change the environment and reveal information about it. Opening a file, querying a service, running a test, and inspecting a spreadsheet are all actions that update the agent's picture of what is true. An action can also be an abstract sequence that must be decomposed into lower-level goals.

Complex tasks contain a hierarchy of these loops. A task such as reviewing a company's software subscriptions and recommending which vendor contracts to renew, renegotiate, or cancel is too abstract to execute directly. The agent must first decompose it:

\begin{enumerate}
  \item Locate the contracts and the finance workbook that tracks spend.
  \item Determine the decision criteria: the budget target, usage thresholds, and notice periods for cancellation.
  \item Extract the active subscriptions and their renewal dates.
  \item Pull usage data for each tool from its admin dashboard.
  \item Normalize the figures so tools are comparable, such as monthly versus annual billing and per-seat versus flat pricing.
  \item Match spend against usage and flag tools that are underused or redundant.
  \item Investigate edge cases, such as a rarely used tool that supports a critical integration.
  \item Produce the renew, renegotiate, or cancel recommendation.
\end{enumerate}

\begin{figure}[H]
\centering
\includegraphics[width=\linewidth]{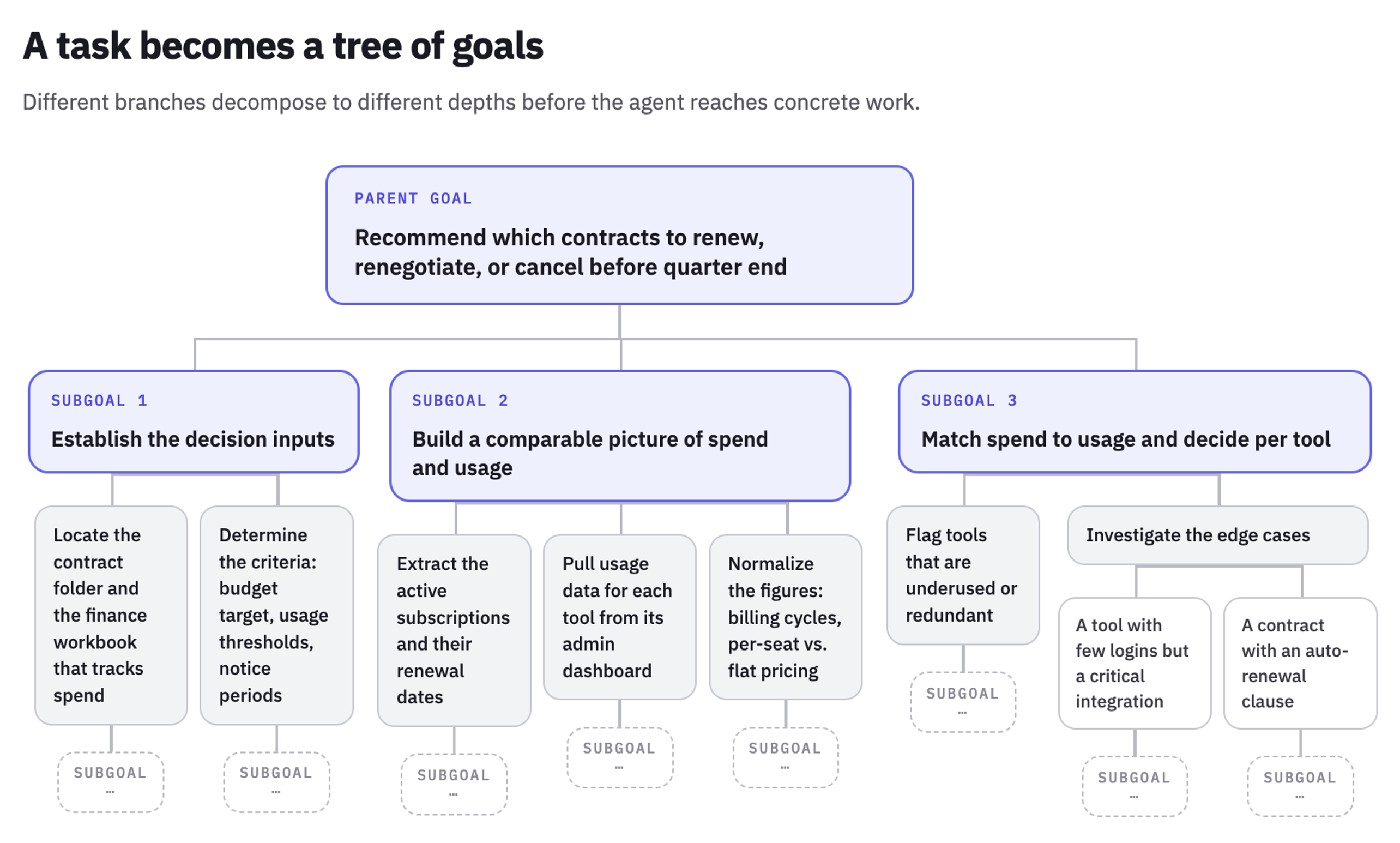}
\caption{A task becomes a tree of goals. The subscription-review task decomposes into branching subgoals, and different branches decompose to different depths before the agent reaches concrete work.}
\label{fig:goaltree}
\end{figure}

Each step is a subgoal that runs its own instance of the loop, and most decompose further before reaching concrete tool calls: pulling usage data splits into finding each tool's dashboard, exporting activity reports, handling tools that offer no export, and merging the results into one comparable sheet. A modest office task expands into a tree several levels deep and dozens of loops wide (Figure~\ref{fig:goaltree}). Some subgoals are sequential, while others form parallel branches whose results must later be synthesized.

Decomposition accounts for only part of the difficulty. The agent must fold low-level observations into one coherent working state and understand what each observation means both for the immediate subgoal and for the objectives above it. This bookkeeping can fail in characteristic ways:

\begin{itemize}
  \item \textbf{A locally sensible goal can be globally wrong.} Converting every contract to a monthly cost simplifies the comparison, but hides the fact that one annual contract auto-renews next week, which is the deadline the task is about.
  \item \textbf{A relevant fact can be lost between levels.} While pulling usage data, the agent notes that most activity runs through service accounts, then carries only the human logins into the comparison, and a heavily used tool looks dormant.
  \item \textbf{A subtask can succeed while undermining a parent requirement.} Cancelling a cheap, barely used tool closes the budget gap while silently breaking an integration that a retained tool depends on.
  \item \textbf{An intermediate result can be coherent yet incomplete.} A spend list built from the finance workbook is internally consistent, but nothing in the list reveals the subscriptions expensed on individual cards that never reached the workbook.
\end{itemize}

\begin{figure}[H]
\centering
\includegraphics[width=0.85\linewidth]{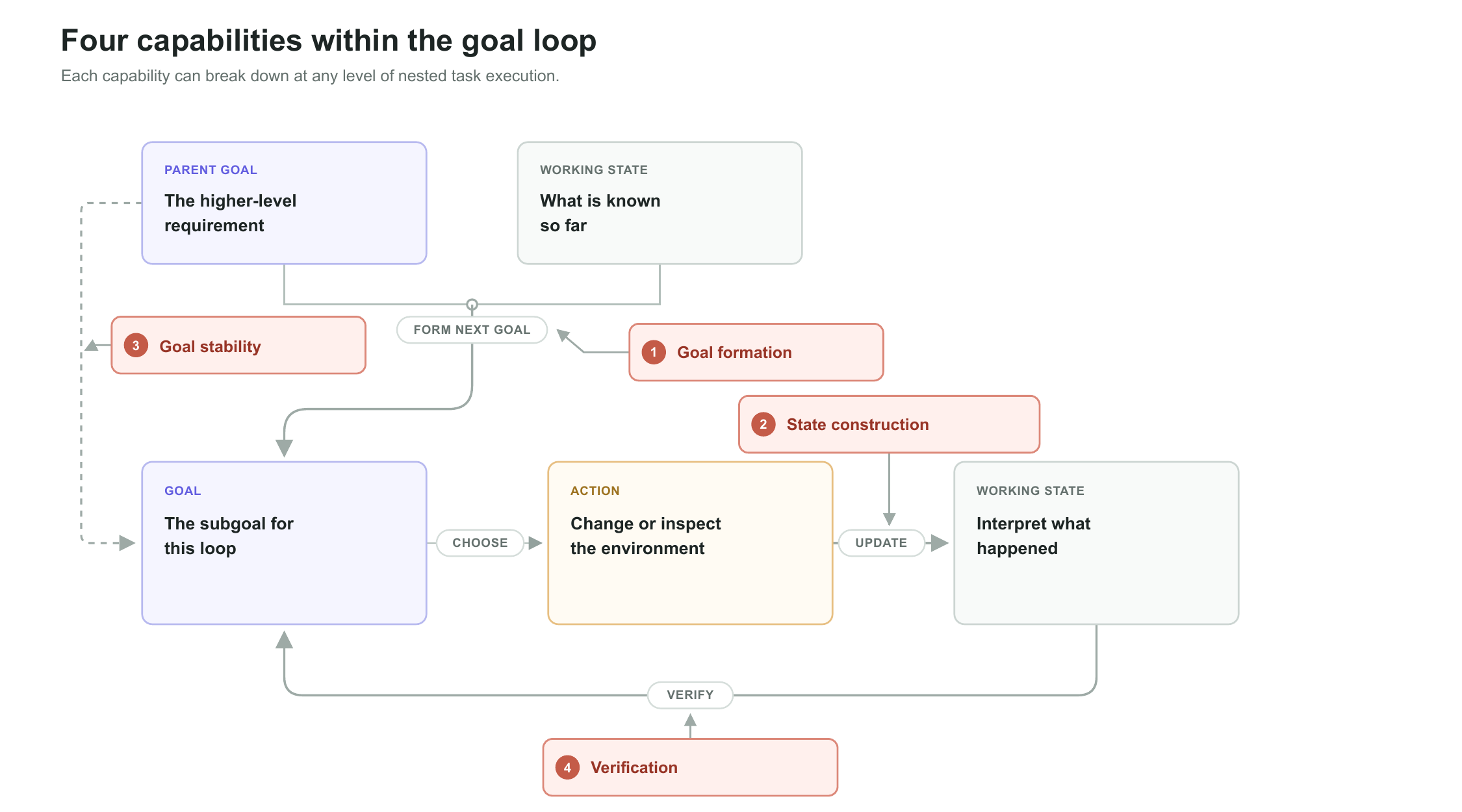}
\caption{Four capabilities within the goal loop: goal formation, state construction, goal stability, and verification. Each can break down at any level of nested task execution.}
\label{fig:failurepoints}
\end{figure}

These examples illustrate breakdowns in the four capabilities through which we operationalize GDE (Figure~\ref{fig:failurepoints}). Table~\ref{tab:capabilities} states the general form of each breakdown.

\begin{table}[H]
\centering
\caption{The four capabilities and examples of how each can break down.}
\label{tab:capabilities}
\begin{tabular}{p{0.26\linewidth}p{0.66\linewidth}}
\toprule
Capability & How it breaks \\
\midrule
Goal formation & The model turns the parent objective and working state into the wrong local target: a simpler proxy, the wrong scope, or a locally plausible goal that is wrong in the broader context. \\
\addlinespace
State construction & The model misses, discards, or fails to combine information needed to understand the environment in relation to the goal. \\
\addlinespace
Goal stability & A local obstacle or complicated subgoal displaces a higher-level goal the model had originally represented correctly. \\
\addlinespace
Verification & The model skips verification or relies on a partial result or proxy, at the level of the overall task or an intermediate subgoal. \\
\bottomrule
\end{tabular}
\end{table}

The goal loop is functionally aligned with TOTE \citep{miller1960plans}: both describe recursive, hierarchical cycles that compare an intended condition with represented current state, operate to reduce the discrepancy, and test again. Our use of the structure differs in purpose. We infer its elements from the behavior of learned tool-using agents and use the four capabilities to compare trajectories across training conditions and domains.

\section{Task Structure and the Transfer Hypothesis}
\label{sec:taskdesign}

Our transfer hypothesis is that tasks from different domains can place similar demands on GDE. LHMTA tasks were designed to reproduce the complexity of realistic work, and four task demands were targeted:

\begin{itemize}
  \item \textbf{Deep decomposition:} high-level objectives require breaking into many parallel and sequential subtasks.
  \item \textbf{Parallel investigation and synthesis:} evidence needed for a decision is distributed across files, services, and tools.
  \item \textbf{Entangled constraints:} high-level requirements interact, making local decisions challenging.
  \item \textbf{Long dependent chains:} later work depends on earlier extraction, interpretation, and modification steps being correct.
\end{itemize}

\begin{figure}[H]
\centering
\includegraphics[width=0.85\linewidth]{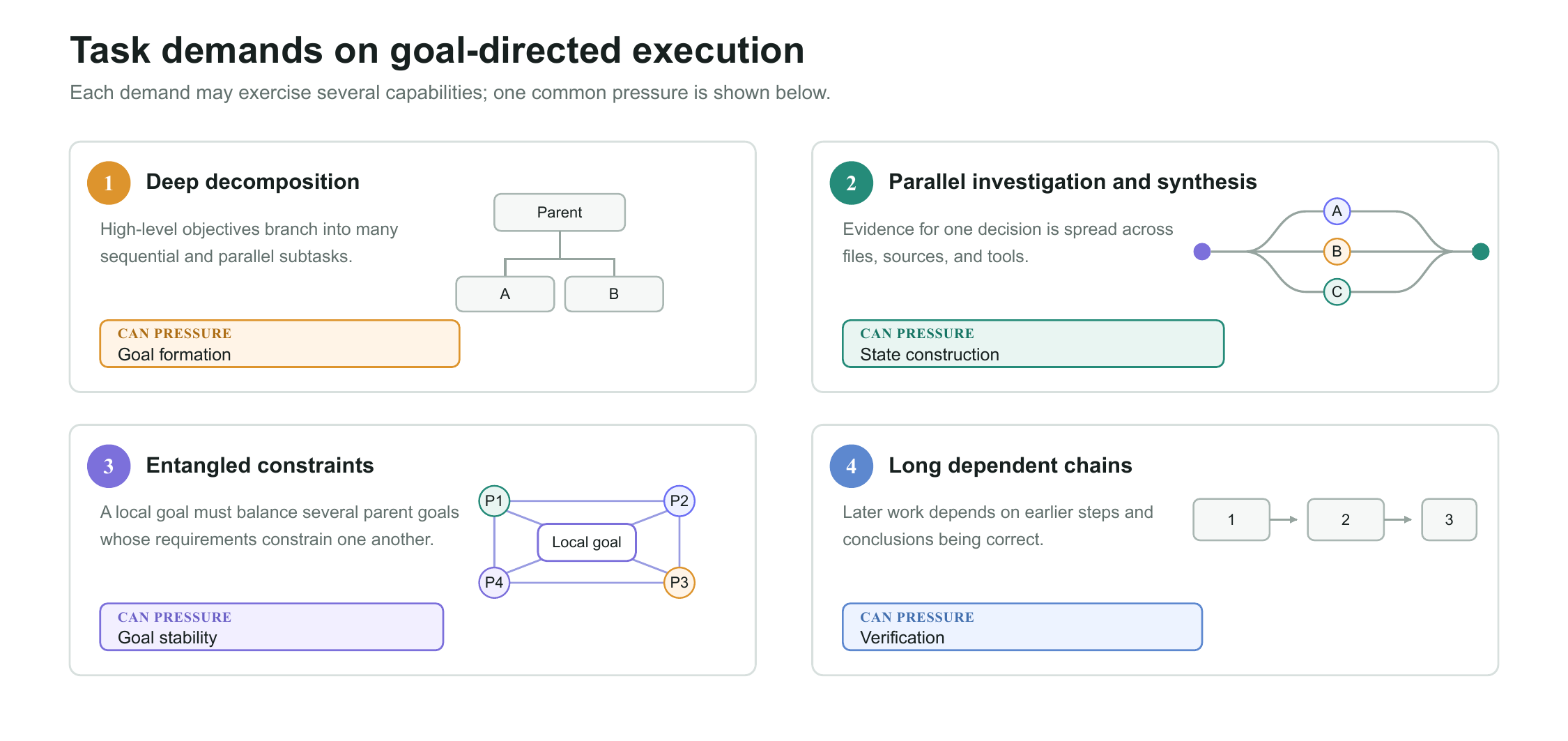}
\caption{Four task demands in LHMTA: deep decomposition, parallel investigation and synthesis, entangled constraints, and long dependent chains. Each may exercise several aspects of GDE; the figure highlights one common pressure for each.}
\label{fig:taskdifficulty}
\end{figure}

These demands may exercise several capabilities at once (Figure~\ref{fig:taskdifficulty}). For example, deep decomposition repeatedly requires goal formation; parallel investigation requires state construction across distributed evidence; entangled constraints place pressure on goal stability; and long dependent chains make intermediate verification consequential because an unchecked error can contaminate every later step. Related cross-domain RL results associate observation richness and policy trajectory length with generalization \citep{liu2026generalizationtax}, but do not isolate the recursive structures considered here. These relationships motivate the transfer hypothesis, but the present experiment did not isolate their causal effects.

The breadth of tools and topics exposed the model to varied instances of these demands. A shortcut tied to one spreadsheet library, tool schema, or output template helps on only a small part of the collection, whereas improvements in relating actions to goals and environment state can remain useful across tasks.

To illustrate the scale of one realistic task, we reconstructed a successful LHMTA trajectory as a recursive task graph. Each node records a goal, the action taken, and the resulting state; child nodes represent decomposed subtasks, parallel branches represent work that can proceed independently, and follow-on nodes combine their results. An evidence catalog links the modeled nodes to the corresponding trajectory events.

The reconstructed hierarchy contains:

\begin{center}
\begin{tabular}{lr}
\toprule
Structural measure & Value \\
\midrule
Task-level goal-loop nodes & 53 \\
Maximum nesting & 10 task levels \\
Leaf subtasks & 29 \\
Parallel fan-out points & 3 \\
Widest parallel fan-out & 5 subtasks \\
Sequential follow-on or synthesis nodes & 21 \\
Tool calls in the source trajectory & 42 \\
Normalized trajectory events & 120 \\
\bottomrule
\end{tabular}
\end{center}

The complete reconstruction, together with its evidence catalog, is available as an \href{https://logan-surge.github.io/lhmta-task-decomposition/}{interactive task decomposition}. It illustrates the structural complexity of a single task but was not used in the aggregate analyses of the dataset or training effects.

\section{Experimental Setup and Results}
\label{sec:setup}

The experiment compared a source domain of long-horizon office and general tool-use workflows with a target domain of software engineering. The training collection contained no software-engineering tasks, graders, or benchmark feedback. The two domains differed in subject matter and tools but shared the need to manage long, stateful tasks.

\subsection{Training environment}
\label{sec:environment}

LHMTA consists of realistic professional workflows exposed through Model Context Protocol servers. Its 27 task categories cover document, spreadsheet, and slide manipulation; search and retrieval; file management; scheduling and calendars; browser automation; planning; and tool use across office services. Individual tasks frequently require the agent to coordinate several services within one trajectory. The collection contains no software-engineering tasks.

Each task has a deterministic Python grader that evaluates the final environment state against several criteria. A task receives a strict pass only when all criteria are satisfied, while the criterion-level structure provides a partial-credit training signal. Successful trajectories typically contain 30 to 40 tool-calling turns and 80,000 to 100,000 tokens.

Task authors targeted realistic work that strong models could not reliably complete and the collection as a whole was monitored for diversity of tools, task types, and failure modes. The dataset snapshot used for this experiment contained 403 tasks: 363 for training and 40 reserved for in-distribution evaluation.

\subsection{Training}
\label{sec:training}

The base model was Qwen3.5-122B-A10B, an open-weight mixture-of-experts model with 122 billion total and approximately 10 billion active parameters. We adapted it using LoRA on the attention and MLP projections \citep{hu2022lora}.

Training proceeded in two stages. First, the model received a supervised warm-up on 3,000 trajectories generated by Kimi K2.6 on LHMTA tasks and rejection-sampled to retain trajectories scoring above 0.9. The warm-up made reinforcement learning feasible: the base policy rarely completed enough of a long-horizon task to receive a sparse pass/fail reward.

Second, we trained on all 363 tasks using the GSPO sequence-level estimator \citep{zheng2025gspo}. We sampled eight rollouts per prompt, and each full trajectory received a dense reward equal to the fraction of grader criteria it satisfied. Reward was assigned at the trajectory level rather than to individual tool-call prefixes. Evaluation used each benchmark's native scoring rather than the dense training reward.

The SFT teacher generated trajectories only on LHMTA tasks, and the external benchmarks provided no training tasks, graders, checkpoint-selection signal, hyperparameter-tuning signal, or reward-design signal.

\subsection{Evaluation}
\label{sec:evaluation}

We compared the base and trained checkpoints on the in-distribution holdout and on external benchmarks. Toolathlon measures long-horizon orchestration across heterogeneous tools \citep{li2025toolathlon}, BFCL-V4 evaluates function calling \citep{patil2025bfcl}, and SWE-Bench Pro requires agents to resolve issues in realistic repositories \citep{deng2025swebenchpro}. All results are pass@1 under greedy decoding.

Table~\ref{tab:toolresults} reports the in-distribution holdout and the external tool-use benchmarks. The holdout result establishes improvement in the target environment, and Toolathlon and BFCL-V4 show that some gains extend beyond LHMTA's particular tasks, graders, and tool conventions.

\begin{table}[H]
\centering
\caption{In-distribution and external tool-use results for the base and trained checkpoints.}
\label{tab:toolresults}
\begin{tabular}{llrrr}
\toprule
Benchmark & Domain relationship & Base pass@1 & Trained pass@1 & Change \\
\midrule
LHMTA holdout & In-distribution & 10.0\% & 27.5\% & +17.5pp \\
Toolathlon & External tool use & 22.2\% & 31.8\% & +9.6pp \\
BFCL-V4 & External function calling & 55.7\% & 59.2\% & +3.5pp \\
\bottomrule
\end{tabular}
\end{table}

Table~\ref{tab:sweresults} reports the cross-domain result examined in this paper: SWE-Bench Pro improved even though the training collection contained no software-engineering tasks. Benchmark transfer alone does not identify what changed in the model's behavior. Section~\ref{sec:analysis} examines whether the transfer is consistent with stronger GDE.

\begin{table}[H]
\centering
\caption{Cross-domain SWE-Bench Pro performance for the same checkpoints.}
\label{tab:sweresults}
\begin{tabular}{lrrr}
\toprule
Benchmark & Base pass@1 & Trained pass@1 & Change \\
\midrule
SWE-Bench Pro & 20.5\% & 26.3\% & +5.8pp \\
\bottomrule
\end{tabular}
\end{table}

\section{Behavioral Analysis of Cross-Domain Transfer}
\label{sec:analysis}

\subsection{Method}
\label{sec:method}

We analyzed all 103 tasks on which the base model failed and the trained model passed: 12 LHMTA holdout cases, 77 SWE-Bench Pro cases, and 14 Toolathlon cases. Each case compared a base and a trained trajectory for the same task under the same prompt, environment, scaffold, and evaluation protocol, giving 206 trajectories in total. We selected this outcome-conditioned subset to study how newly successful behavior manifested. It does not estimate the prevalence of each capability across the full evaluation distribution.

We used Claude Opus 4.8 as a root-cause-analysis agent. For each scored output failure, the agent traced backward through the failed trajectory to identify the earliest evidence-supported point at which it diverged from a successful approach. It examined task requirements, evaluator evidence, tool calls and results, produced artifacts, edits, tests, and model reasoning. Each report cited the specific trajectory events and artifacts supporting its causal account, allowing us to audit each report against the source material. The successful trajectory supplied contrast evidence for a task-correct path but was not treated as automatic ground truth.

The reports accelerated the analysis by locating and documenting candidate divergence points; they did not independently validate the framework. We manually reviewed the reports against the corresponding raw trajectories, checking claims and quotations against the original reasoning messages, tool calls, environment responses, edits, and tests before using any case as an example. Through iterative comparison across cases, we organized recurring behavioral differences into four capabilities: goal formation, state construction, goal stability, and verification. These capabilities form an interpretive framework rather than mutually exclusive labels.

We supplemented the case studies with deterministic measures over matched SWE-Bench Pro trajectories. These measures cover retrieval repetition and breadth, contact with files changed by the reference patch, patch size, and the presence and timing of formal tests. They provide broader but indirect signals of behavioral change.

\subsection{Four capabilities in paired trajectories}
\label{sec:capabilities}

Table~\ref{tab:manifestations} summarizes how each capability appears in the paired examples. The surface failures differ by domain, but each pair reflects the same broken relationship among the current goal, higher-level requirements, and environment evidence.

\begin{table}[H]
\centering
\caption{The four capabilities across paired general tool-use and software-engineering cases.}
\label{tab:manifestations}
\begin{tabular}{p{0.24\linewidth}p{0.34\linewidth}p{0.34\linewidth}}
\toprule
Capability & General tool-use manifestation & Software-engineering manifestation \\
\midrule
Goal formation & Chose a locally plausible calculation that conflicted with the meaning of the task & Reimplemented behavior despite existing helpers that already encoded the required semantics \\
\addlinespace
State construction & Discarded formula-backed values needed by the parent task & Failed to connect the new requirement to the repository's existing validation structure \\
\addlinespace
Goal stability & Reintroduced a course that violated a previously identified course-level constraint & Changed correct parser behavior to satisfy a stale local test \\
\addlinespace
Verification & Treated a bounded extraction as a complete catalog & Checked new exports but left downstream consumers calling removed APIs \\
\bottomrule
\end{tabular}
\end{table}

\paragraph{Goal formation.}
Every loop requires converting a prompt or parent action into a concrete target grounded in both the higher-level objective and the environment. A common failure mode is choosing a locally plausible target that is wrong in the broader context.

\textbf{SWE-Bench Pro: NodeBB selected fields.} The task required NodeBB database callers to request selected fields from stored objects. Both models found existing helpers and tests that already implemented the required selected-field semantics, including preserving one result per key and representing requested-but-missing fields as \texttt{null}. The correct local goal was to route the public methods through these helpers. The base model instead reimplemented field filtering inside backend paths and changed missing-object behavior. The trained model reused the established helpers, including in the Postgres path.

\begin{figure}[H]
\centering
\includegraphics[width=0.9\linewidth]{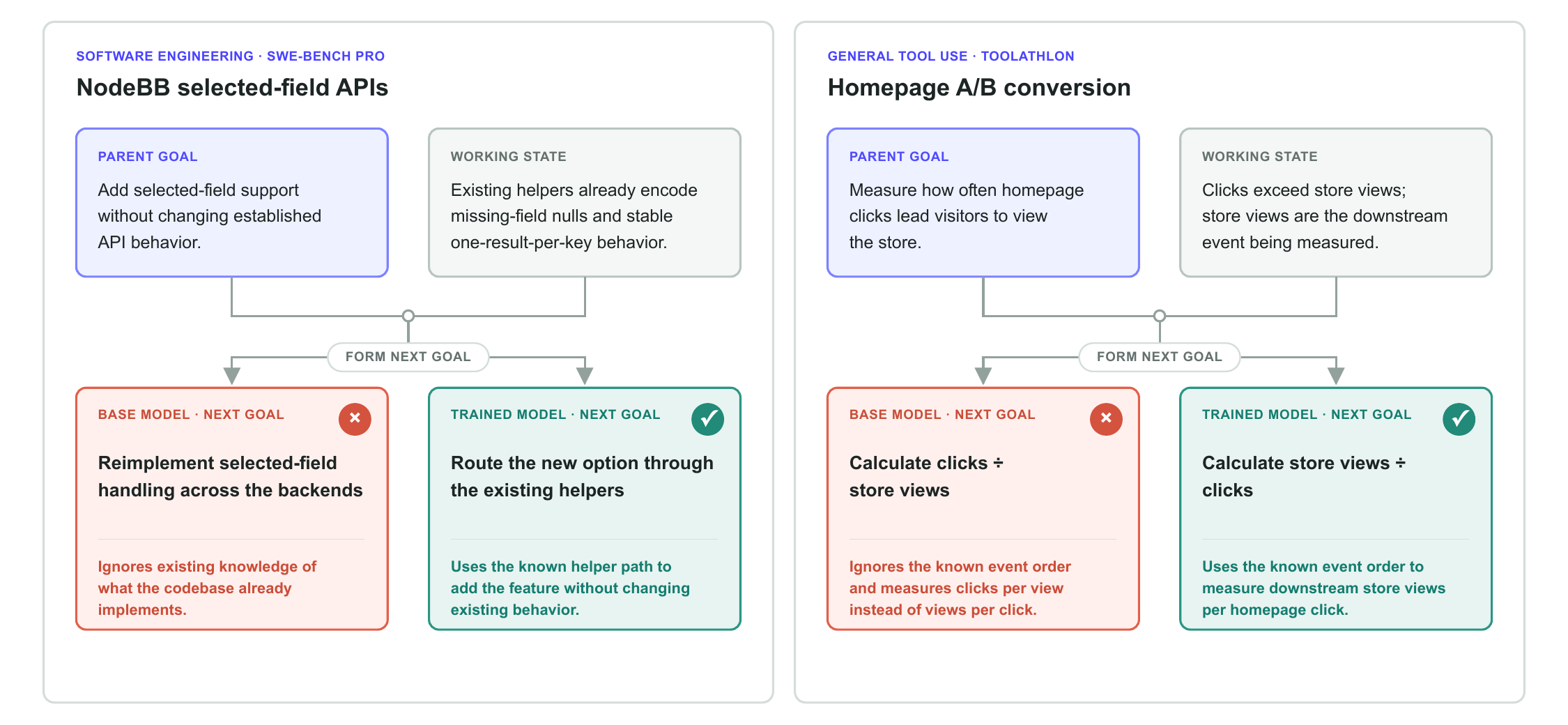}
\caption{Goal formation across domains. From the same parent goal and working state, the base model forms a locally plausible but incorrect local goal, while the trained model forms the correct one.}
\label{fig:goalcontrast}
\end{figure}

\textbf{Toolathlon: A/B conversion rates.} The task required calculating how often homepage clicks led to store views. The base model defined conversion as \texttt{clicks / store\_views} before inspecting any values. When this produced rates above 100 percent, it noted that the result was strange but continued. The trained model interpreted the columns in relation to the parent objective and used \texttt{store\_views / clicks}. In both domains, the difference lay in how environment information shaped the next local goal (Figure~\ref{fig:goalcontrast}).

\paragraph{State construction.}
The agent must decide what to inspect and integrate its observations into a working state that supports both local and higher-level decisions. Relevant evidence can be missed, misinterpreted, or understood only within a low-level subtask.

\textbf{SWE-Bench Pro: Ansible collection-name validation.} The task required collection-name validation to reject reserved Python keywords. The base model inspected the relevant imports and validation structure, then introduced a helper that called \texttt{keyword.iskeyword} even though the module imported \texttt{iskeyword} directly. The resulting \texttt{NameError} caused 19 tests to fail. When a self-test exposed the problem, the model changed the test setup and left its implementation untouched. The trained model connected the requirement to the repository's central \texttt{AnsibleCollectionRef.is\_valid\_collection\_name} validator and reused that path.

\begin{figure}[H]
\centering
\includegraphics[width=0.9\linewidth]{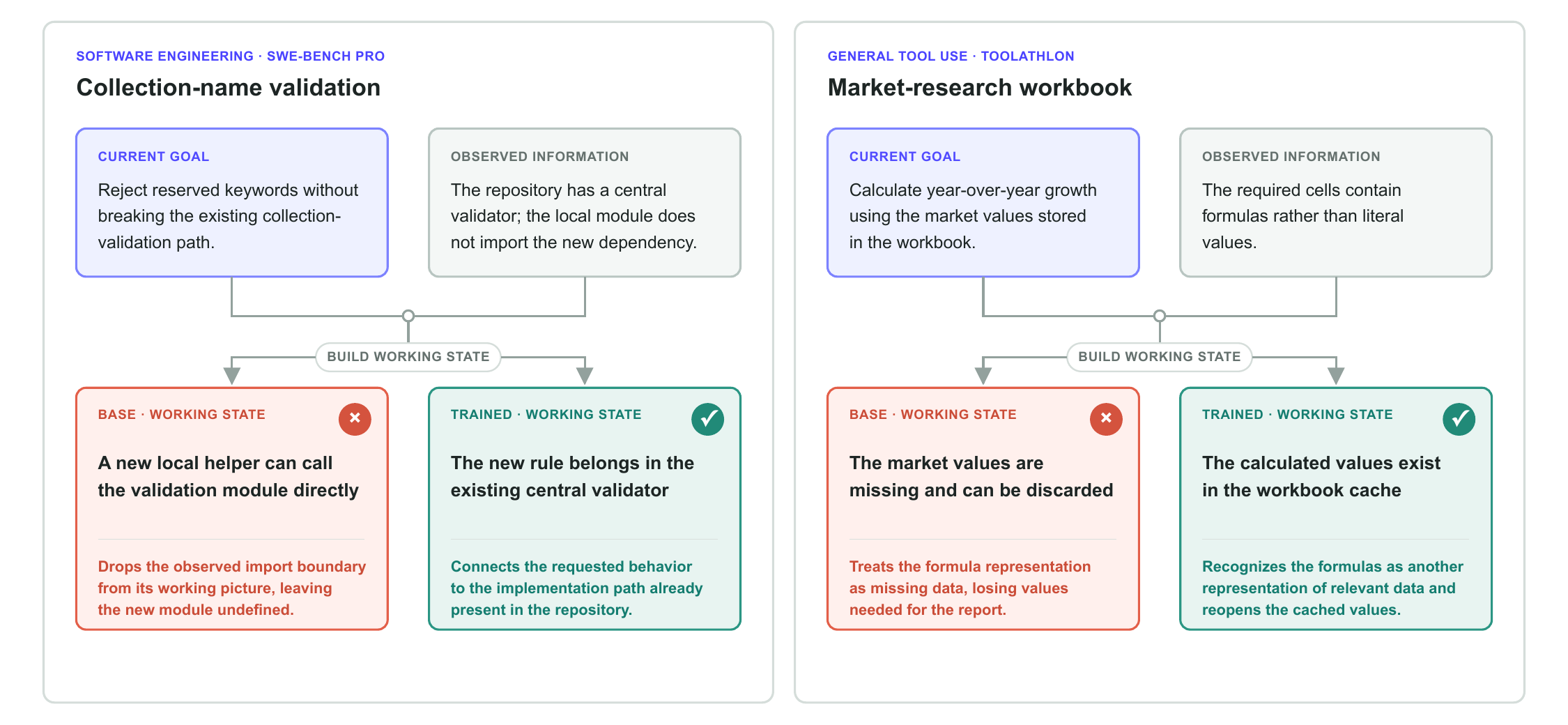}
\caption{State construction across domains. The base model drops or misrepresents a relevant observation; the trained model incorporates it into the working state used by the parent task.}
\label{fig:statecontrast}
\end{figure}

\textbf{Toolathlon: formula-backed workbook values.} A market-research task required year-over-year growth computed from formula-backed cells. The base model encountered formula strings during extraction and discarded the entries as non-numeric, even though their calculated values were required by the report. The trained model reopened the workbook using cached formula values and retained the calculated data. Both models encountered the same tool friction, but only the trained model incorporated the observation into the state needed by the parent task (Figure~\ref{fig:statecontrast}).

\paragraph{Goal stability.}
A difficult local subtask can displace a requirement the model previously represented correctly.

\textbf{SWE-Bench Pro: PowerShell CLIXML parsing.} The parser had to decode escaped characters while preserving control characters such as carriage returns and line feeds. The base model initially implemented the requirement correctly. It then encountered an existing test expecting a trailing CRLF to be stripped and changed the implementation to satisfy the stale behavior. The trained model interpreted the test relative to the requested behavior and retained the CRLF-preserving implementation.

\begin{figure}[H]
\centering
\includegraphics[width=0.9\linewidth]{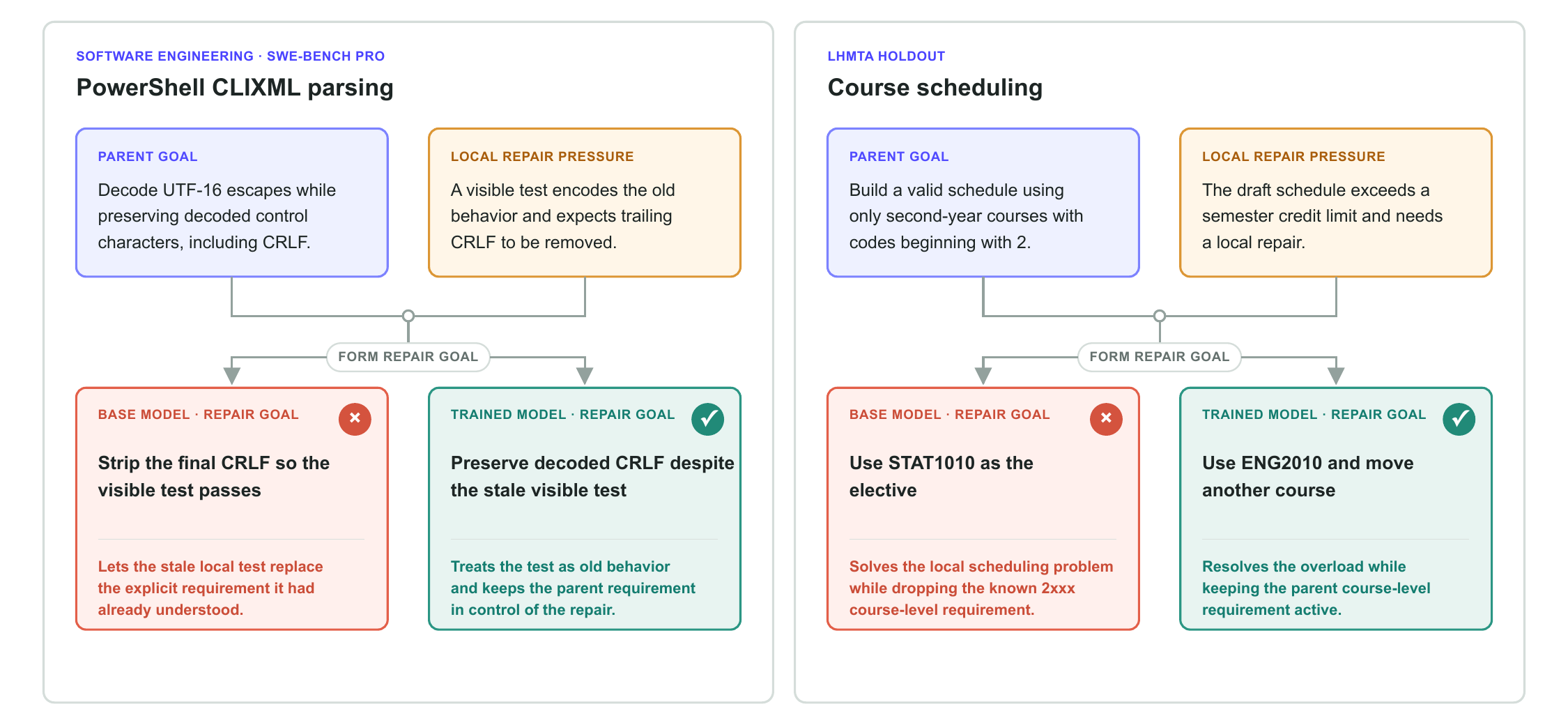}
\caption{Goal stability across domains. A parent-goal constraint enters a local repair loop; the base model displaces it, while the trained model preserves it.}
\label{fig:stabilitycontrast}
\end{figure}

\textbf{LHMTA holdout: course scheduling.} Every selected course had to be at the second-year \texttt{2xxx} level. The base model recorded the constraint and correctly identified \texttt{STAT1010} as invalid. Later, while repairing an overloaded semester, it selected \texttt{STAT1010} as an elective and claimed that all courses satisfied the rule. The trained model preserved the constraint, selected \texttt{ENG2010}, and moved another course to resolve the overload. In both pairs, local repair pressure displaced a known parent criterion in the base trajectory but not in the trained one (Figure~\ref{fig:stabilitycontrast}).

\paragraph{Verification.}
Verification closes goal loops at both task and subtask boundaries. The model must determine what evidence would establish that the intended environment state has been reached.

\textbf{SWE-Bench Pro: Element Web \texttt{Pill} refactor.} The task required converting a \texttt{Pill} class component into a function and exporting helpers that had been static methods. The base model checked that the new helpers were exported and imported. Its validation stopped before searching for consumers of the removed static APIs, leaving downstream code that still called \texttt{Pill.roomNotifPos} and \texttt{Pill.roomNotifLen}. The trained model searched for stale usages, migrated the call sites, and ran a targeted TypeScript check.

\begin{figure}[H]
\centering
\includegraphics[width=0.9\linewidth]{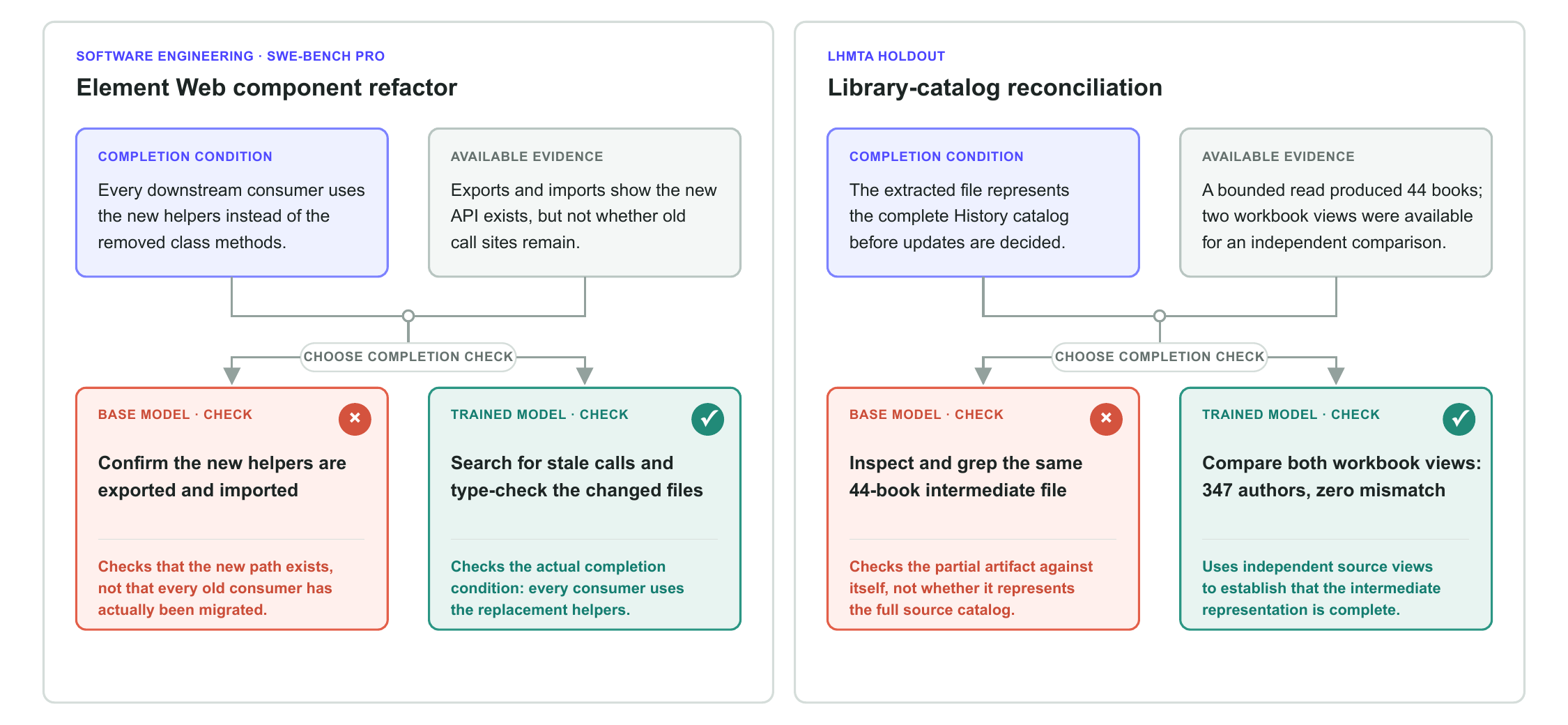}
\caption{Verification across domains. The base model relies on proxy or self-referential evidence; the trained model checks the relevant claim against the environment.}
\label{fig:verifycontrast}
\end{figure}

\textbf{LHMTA holdout: catalog reconciliation.} A library task required reconciling History books against records distributed across several files. The base model created a 44-book intermediate file from a bounded spreadsheet read and inspected that file, but never checked whether the read covered the full catalog. The trained model compared two workbook views and confirmed that each contained the same 347 History authors before using the extraction. The base model's checks established that an artifact existed and was internally coherent; the trained model's checks tested the relevant claim against the environment (Figure~\ref{fig:verifycontrast}).

\subsection{Aggregate behavioral evidence}
\label{sec:aggregate}

The case studies show how individual tasks diverge; the aggregate SWE-Bench Pro measures examine whether related behavioral signals appear more broadly (Table~\ref{tab:metrics}).

\begin{table}[H]
\centering
\caption{Behavioral metrics from SWE-Bench Pro trajectories, computed over 731 paired tasks.}
\label{tab:metrics}
\begin{tabular}{p{0.38\linewidth}rrp{0.3\linewidth}}
\toprule
Behavioral signal & Base & Trained & Interpretation \\
\midrule
Mean retrieval calls per run & 25.6 & 23.9 & Similar retrieval effort \\
Mean distinct retrieved information & 7,985 & 9,019 & More unique retrieved content \\
Repeated retrieved information & 22.5\% & 14.3\% & Less repeated investigation \\
Mean reference-patch files touched & 2.69 & 3.08 & Greater reference-file overlap \\
Mean added lines & 415.5 & 111.7 & Smaller patches \\
Runs containing a formal test & 37.5\% & 73.3\% & More frequent verification \\
Mean first-test position (tested runs) & 70.8\% & 62.0\% & Earlier verification \\
Runs with passing test followed by edit & 5.9\% & 12.9\% & More intermediate verification \\
\bottomrule
\end{tabular}
\end{table}

The metrics are defined as follows:

\begin{itemize}
  \item \textbf{Mean retrieval calls per run.} Read-only inspection commands, such as ls, grep, cat, and read-only git subcommands, that neither modify the environment nor redirect output.
  \item \textbf{Mean distinct retrieved information.} Mean number of unique 12-token spans in retrieval-call outputs per run.
  \item \textbf{Repeated retrieved information.} The share of spans already seen earlier in the same run.
  \item \textbf{Mean reference-patch files touched.} Mean number of files per run changed by both the agent's final patch and the task's reference patch.
  \item \textbf{Mean added lines.} Added lines in the agent's final patch.
  \item \textbf{Runs containing a formal test.} Runs with at least one command invoking a standard test runner such as pytest or \texttt{go test}.
  \item \textbf{Mean first-test position.} How far through a run's commands the first test appears, from zero to one, among tested runs.
  \item \textbf{Runs with a passing test followed by editing.} Runs in which a passing test is followed by a further file edit.
\end{itemize}

Appendix~\ref{app:metrics} gives the exact classification patterns and parsing rules behind each metric.

The trained model made slightly fewer retrieval calls while covering more distinct information and repeating less content, a pattern consistent with a better-maintained picture of what has already been learned and what remains unknown. It also touched more of the files changed by the reference patch while adding approximately one quarter as many lines, consistent with more targeted implementation.

Testing provides the clearest aggregate change. The trained model ran formal tests in nearly twice as many trajectories and began testing earlier. Passing tests were also more likely to be followed by further editing, suggesting that verification frequently informed intermediate work rather than serving only as a final check.

These measures provide convergent behavioral signals rather than direct measurements of the four capabilities. Reference patches do not uniquely specify a valid implementation strategy, and smaller patches are not inherently better. The available telemetry also lacks a clean aggregate proxy for goal stability, so evidence for that capability comes from the paired cases.

\section{What Long-Horizon Post-Training Teaches}
\label{sec:discussion}

The results support a behavioral interpretation in which post-training changed how the model organized work over long tasks. The training collection supplied no software-engineering tasks, repository conventions, or code-level solutions. Yet the same differences in goal formation, state construction, goal stability, and verification appeared in both office workflows and software repositories. GDE provides a common description of these changes without requiring the source and target tasks to share subject matter.

One plausible account is that the training improved how reliably the model deployed knowledge acquired during pretraining. Prior work has argued that pretrained models already contain much of the knowledge required for downstream tasks and that instruction tuning improves their ability to apply it in response to user intent \citep{zhou2023lima, ouyang2022instructions}. LHMTA may have provided repeated practice applying available knowledge while pursuing complex goals in unfamiliar, stateful environments. The software result is informative because the domain mismatch makes direct acquisition of repository-specific knowledge from the training set implausible.

The causal explanation remains a hypothesis. The experiment did not isolate deep decomposition, parallel synthesis, entangled constraints, or dependent chains, and it did not establish which of these demands contributed to transfer. Cross-domain evaluation can reveal when training effects extend beyond surface workflows, while the four capabilities provide a vocabulary for more controlled questions: which task properties strengthen each capability, which capabilities transfer together, and whether the same patterns replicate across models, data mixtures, and training methods.

Long-horizon agentic data is costly to author and train on; successful LHMTA trajectories often span 80,000 to 100,000 tokens. This raises the value of understanding what kinds of behavior a training experience exercises. The present results motivate that question but do not yet provide a recipe for engineering particular task demands.

\section{Limitations}
\label{sec:limitations}

\paragraph{Single model and training run.} All results come from one base model and one post-training run. We have not established reproducibility across seeds, scales, model families, training recipes, or data mixtures.

\paragraph{Exploratory behavioral interpretation.} The four capabilities were refined iteratively through qualitative analysis of trajectories. They describe observable relationships among goals, actions, environment evidence, and outcomes; they do not identify literal internal goals, state registers, or control loops and do not constitute an exhaustive taxonomy.

\paragraph{Outcome-conditioned cases and analyst judgment.} The analysis focused on tasks where the base model failed and the trained model passed, and the cases shown were selected for explanatory clarity rather than by random sampling. They illustrate how improvement manifested but do not estimate the prevalence of each capability. Reconstructing goals and causal chains from long trajectories also involves judgment, and individual failures can have multiple causes.

\paragraph{No causal or domain-matched counterfactual.} We did not ablate the proposed task demands or compare LHMTA with an equal budget of software-engineering training data. The experiment therefore does not establish which properties caused transfer or whether non-software training was the most efficient way to improve SWE-Bench Pro. Broader changes in exploration, persistence, instruction following, or tool use remain possible explanations for part of the gain.

\paragraph{Proxy measures.} Retrieval overlap, reference-file coverage, patch size, and testing commands are indirect behavioral signals. Reference patches are not unique ground truth.

\section{Conclusion}
\label{sec:conclusion}

Post-training Qwen3.5-122B-A10B on 363 long-horizon multi-tool tasks produced behavior that remained useful outside the training domain: SWE-Bench Pro improved even though the training collection contained no software-engineering tasks. Paired trajectories showed recurring improvements in goal formation, state construction, goal stability, and verification, while aggregate measures provided broader evidence of changes in investigation, implementation, and testing. GDE organizes these differences as a behavioral account of how the trained model pursued long tasks.

The findings motivate studying post-training data as experience that exercises behavioral capabilities, not only as examples containing particular topics, tools, or answers. Establishing the causal relationship will require controlled comparisons across task demands, domains, models, and training methods. A science of post-training should explain not only whether a dataset improves a benchmark, but what kinds of behavioral capabilities the training experience develops and where those capabilities transfer.

\bibliographystyle{plainnat}
\bibliography{references}

\appendix

\section{Exact Metric Definitions}
\label{app:metrics}

This appendix gives the exact classification rules behind the metrics in Table~\ref{tab:metrics}. All patterns are regular expressions applied case-insensitively to individual assistant shell commands. Long patterns are wrapped across lines for readability; the implemented patterns contain no whitespace at the wrap points.

\subsection{Retrieval calls}
\label{app:retrieval}

A command is classified as a retrieval call in three steps. Occurrences of \verb!>/dev/null! are ignored; any other output redirection disqualifies the command. A command matching the exclusion pattern below is disqualified. The remaining commands qualify if they match the inclusion pattern.

Inclusion pattern:

\begin{verbatim}
(?:^|[;&|()]\s*)(?:
  ls|find|fd|rg|grep|egrep|fgrep|cat|head|tail|wc|stat|file|tree|pwd|
  readlink|realpath|which|whereis|type|jq|awk|cut|sort|uniq|
  git\s+(?:show|log|grep|ls-files|rev-parse|branch|remote|blame)
)\b
\end{verbatim}

Exclusion pattern:

\begin{verbatim}
\b(?:rm|mv|cp|mkdir|rmdir|touch|chmod|chown|ln|tee|truncate|patch)\b|
\bgit\s+(?:add|commit|apply|checkout|restore|reset|clean|am|
         cherry-pick|rebase|merge)\b|
\bsed\s+-[^\n;|&]*i\b|\bperl\s+-[^\n;|&]*i\b|
\b(?:python|python3|node|ruby|php)\b|
\b(?:pytest|unittest|jest|vitest|mocha|rspec|go\s+test|cargo\s+test|
    npm\s+test|yarn\s+test|pnpm\s+test)\b|
\b(?:make|cmake|ninja|npm\s+install|yarn\s+install|pnpm\s+install|
    pip\s+install)\b|
<<
\end{verbatim}

Although \texttt{git branch} and \texttt{git remote} can mutate repository state with other arguments, all matched uses in this corpus were read-only: \texttt{git branch} appeared only in listing or inspection forms such as \verb!-a!, \verb!--contains!, and \verb!--show-current!, and \texttt{git remote} appeared only as \verb!git remote -v!. No branch creation, branch deletion, or remote mutation was counted as retrieval.

\subsection{Information spans}
\label{app:spans}

Retrieval-call outputs are processed as follows: ANSI escape sequences are removed, the text is tokenized with the pattern \verb![A-Za-z_][A-Za-z0-9_]*|\d+(?:\.\d+)?|[^\s]! and lowercased, and every overlapping window of 12 consecutive tokens forms one span; an output shorter than 12 tokens forms a single span. Distinct retrieved information is the number of unique spans in a run. Repeated retrieved information is the share of spans that already occurred in an earlier output of the same run.

\subsection{Formal tests and mutations}
\label{app:tests}

A command contains a formal test if it matches any of:

\begin{verbatim}
(?:^|[;&|]\s*|\s)(?:python\d*\s+-m\s+)?pytest(?:\s|$)
(?:^|[;&|]\s*|\s)py\.test(?:\s|$)
\bgo\s+test\b
\bcargo\s+test\b
\b(?:npm|pnpm|yarn)\s+(?:run\s+)?test(?:\s|$|:)
\b(?:npx\s+)?(?:jest|vitest|mocha)(?:\s|$)
\bmake\s+(?:[\w.-]*test[\w.-]*|check)(?:\s|$)
(?:^|[;&|]\s*|\s)(?:tox|nosetests|unittest)(?:\s|$)
\bpython\d*\s+-m\s+unittest\b
\bbazel\s+test\b
\bgradle\w*\s+test\b
\bmvn\w*\s+(?:test|verify)\b
\bdotnet\s+test\b
\brspec(?:\s|$)
\end{verbatim}

A test counts as passing when it returns code zero, is not part of a status-masking pipeline or fallback, and its output contains no explicit failure marker. A command counts as a mutation if it matches any of:

\begin{verbatim}
\bapply_patch\b
\b(?:sed|perl)\b[^\n]*(?:\s-i\b|--in-place)
\bgit\s+(?:apply|checkout|restore|reset|cherry-pick)\b
\b(?:cp|mv|rm|touch)\s+(?![^\n]*(?:/tmp/|/dev/))
\.(?:write_text|write_bytes)\s*\(
\bopen\s*\([^\n]{0,180},\s*['"](?:w|a|x|w\+|a\+)['"]
\b(?:cat|printf|echo)\b[^\n]*(?:>>|>)\s+(?!/tmp/|/dev/)[^\s;&|]+
\btee\s+(?:-a\s+)?(?!/tmp/|/dev/)[^\s;&|]+
\end{verbatim}

First-test position is the ordinal position of the first formal-test command among all of a run's assistant shell commands, normalized to the range zero to one.

\subsection{Patch metrics}
\label{app:patches}

The agent patch is the final submitted diff for a run. The reference patch is the task's gold code patch; the task's separate test patch is not included. File paths are read from \verb!diff --git! headers. Reference-patch files touched is the number of distinct files appearing in both the agent patch and the reference patch. Added lines is the number of lines beginning with \verb!+!, excluding \verb!+++! file headers, summed over all files in the agent patch.

\end{document}